%% file: main.tex
\pdfoutput=1

\documentclass[11pt]{article}

\usepackage{acl}

\usepackage{times}
\usepackage{latexsym}

\usepackage[T1]{fontenc}

\usepackage[utf8]{inputenc}

\usepackage{microtype}

\usepackage{amsmath}
\usepackage{amssymb}
\usepackage{mathrsfs}
\usepackage{graphicx}
\usepackage{multirow}
\usepackage{multicol}
\usepackage{float}
\usepackage{subfigure}
\usepackage{latexsym}
\usepackage{booktabs}
\usepackage{makecell}

\title{Towards Effective Multi-Task Interaction for Entity-Relation \\Extraction: A Unified Framework with Selection Recurrent Network}

\author{An Wang \\
  Tokyo Institute of Technology \\
  \texttt{wang@de.cs.titech.ac.jp} \\\And
  Ao Liu \\
  Tokyo Institute of Technology \\
  \texttt{liu.ao@nlp.c.titech.ac.jp} \\\AND
  Hieu Hanh Le \\
  Tokyo Institute of Technology \\
  \texttt{hanhlh@de.cs.titech.ac.jp} \\\And
  Haruo Yokota \\
  Tokyo Institute of Technology \\
  \texttt{yokota@cs.titech.ac.jp} \\
  }

\begin{document}
\maketitle
\begin{abstract}
Entity-relation extraction aims to jointly solve named entity recognition (NER) and relation extraction (RE). Recent approaches use either one-way sequential information propagation in a pipeline manner or two-way implicit interaction with a shared encoder. However, they still suffer from poor information interaction due to the gap between the different task forms of NER and RE, raising a controversial question whether RE is really beneficial to NER. Motivated by this, we propose a novel and unified cascade framework that combines the advantages of both sequential information propagation and implicit interaction. Meanwhile, it eliminates the gap between the two tasks by reformulating entity-relation extraction as unified span-extraction tasks. Specifically, we propose a selection recurrent network as a shared encoder to encode task-specific independent and shared representations and design two sequential information propagation strategies to realize the sequential information flow between NER and RE. Extensive experiments demonstrate that our approaches can achieve state-of-the-art results on two common benchmarks, ACE05 and SciERC and effectively model the multi-task interaction, which realizes significant mutual benefits of NER and RE.
\end{abstract}

\section{Introduction}

\input{paper/Introduction}

\section{Related Work}

\input{paper/Related_Work}

\input{paper/Method}

\section{Experiments}

\input{paper/Experiments}

\section{Conclusion}

In this paper, we propose a unified cascade framework for entity-relation extraction by reformulating it as three span extraction tasks which unifies the task form and eliminates the gap between NER and RE. To achieve effective multi-task interaction, we propose a selection recurrent network which models implicit interaction among three sub-tasks and
design two prior information fusing strategies to realize explicit sequential information flow between NER and RE. We conduct extensive experiments on two benchmarks to verify the effectiveness of our framework. We also employ ablation studies to explore how different factors impact the final performance and extensive analysis experiments to understand the reason of our improvements. Lastly, our framework successfully averts the weaknesses of previous approaches in modeling the interaction between NER and RE, and realizes the mutual benefits of the two tasks.

\bibliography{reference}
\bibliographystyle{acl_natbib}

\appendix

\section{Appendix}
\label{sec:appendix}
\input{paper/Appendix}

\end{document}

%% file: paper/Introduction.tex
\begin{figure}[t]
    \centering
    \small
    \includegraphics[width=1.1\linewidth]{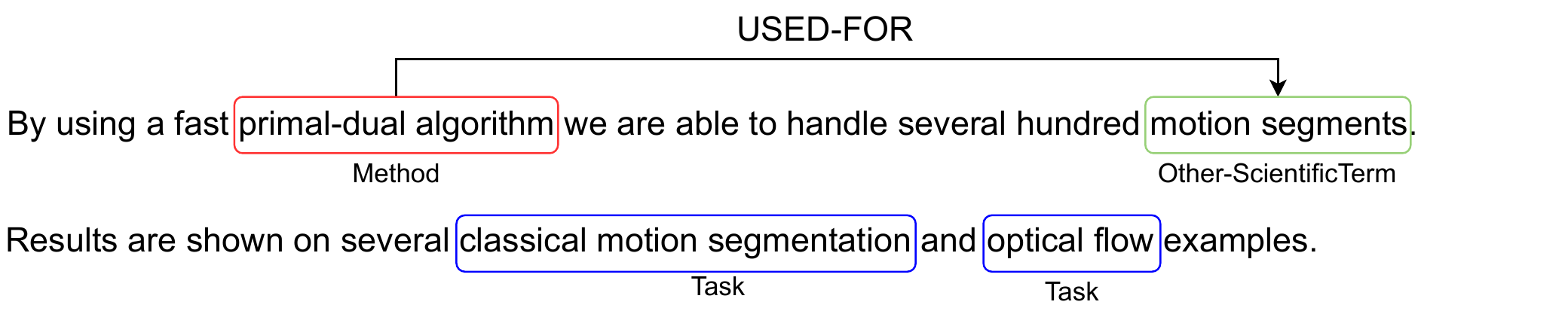}
    \caption{\label{fig:example}Examples of entity-relation extraction. \textit{primal-dual algorithm} and \textit{motion segments} are in-triple entities connected in \textit{USED-FOR} relation. \textit{classical motion segmentation} and  \textit{optical flow} are out-of-triple entities that do not correspond to any relations.}
\end{figure}

Entity-relation extraction is a fundamental problem in Information Extraction (IE). It aims to both identify named entities and extract relations between them. This problem can be decomposed into two sub-tasks: named entity recognition (NER)~\cite{sang2003introduction, zhang2004focused, ratinov2009design} and relation extraction (RE)~\cite{zelenko2003kernel, bunescu2005shortest}. Figure \ref{fig:example} shows an example of entity-relation extraction.

Early works~\cite{sang2003introduction, florian2004statistical} typically adopt a pipeline framework by solving each sub-task with sequential modules. They first recognize all the entities in a sentence and then performs relation classification for each entity pair. These traditional pipeline methods ignore the interaction between NER and RE.

Some recent works adopt a cascading pipeline~\cite{yu2019joint,wei2019novel} which can develop a unified label space by reformulating NER and RE as closer sub-tasks: first extract subject entities, then extract corresponding object entities. A fatal shortcoming of these methods is that they cannot deal with out-of-triple entities (examples shown in Figure \ref{fig:example}) which do not appear in any relation. 

Current approaches aim to jointly solve the two sub-tasks to take advantage of the benefits of their inter-task correlation. Specifically, some of them cast NER and RE as a joint table filling problem \cite{miwa2014modeling,gupta2016table,zhang2017end,tran2019neural,wang2020two,wang2020tplinker,wang2021unire}. This allows NER and RE to be performed in one stage through implicit multi-task interaction, realized by shared feature space. However, these works ignore the natural sequential information propagation between NER and RE. 

Moreover, PURE~\cite{zhong2020frustratingly} revisits pipeline methods and realizes the sequential information flow between NER and RE through inserting typed entity markers into the input text of the RE model, which improves the performance of RE. They empirically find that using a shared encoder to model implicit interaction does not improve but rather impairs the performance of NER. 
We attribute the reason to the fact that they did not consider the contradictory information in the shared features caused by the natural gap between NER and RE. 
In contrast, \citet{yan2021partition} considers this problem and proposes a partition filter network (PFN) to improve the joint table-filling methods by separating task-specific independent information and shared information from the shared representations. They claim the opposite finding that through implicit interaction, RE contributes non-negligibly to NER. However, they still find that RE disturbs the extraction of out-of-triple entities which in essence do not correspond to any relations, resulting in the overall lower NER results compared with \cite{zhong2020frustratingly}. Motivated by these findings, we seek to explore more effective multi-task interaction to alleviate the potential negative impacts of RE and make RE really beneficial to NER.

In this study, we propose a novel and unified cascade framework which eliminates the gap between different task forms.
We treat entity-relation extraction as a multi-turn span extraction problem includes entity extraction, subject extraction and subject-oriented object extraction. We employ a separate task model for each turn.
In this way, we regard NER as entity extraction and decompose RE into subject extraction and subject-oriented object extraction.
For each task, we employ a BERT-based embedder \cite{devlin2018bert} and a shared multi-task encoder to obtain token representations in the sentence. Then we feed these representations into separate task models to perform span extraction.

Towards effective multi-task interaction between NER and RE, we consider two kinds of interactions: implicit interaction and sequential information propagation.
During modeling implicit interaction, to avoid irrelevant information hurting model performance, we propose a \textit{selection recurrent network} (SRN) encoder to learn task-specific independent and shared representation for three tasks. 
For sequential information propagation, we aim to maintain the natural information flow between NER and RE by incorporating the output information from NER into the input of RE. To this end, we propose two strategies of fusing such prior information in our three-task framework: \textit{early fusion} and \textit{late fusion} strategies. 
\textit{early fusion} strategy fuses the prior information into the subsequent task models by inserting subject and entity type markers into the input sentence but leads a a large computational cost due to separate text encoding. While \textit{late fusion} strategy solves this issue by reusing the same encoded representation for all the tasks and propagating the prior information as an intermediate entity/subject embedding.

We evaluate our method on two public datasets: ACE05\cite{walker2006ace} and SciERC\cite{luan2018multi}.
Experimental results demonstrate that our method outperforms previous state-of-the-art models on both benchmarks. In particular, it especially outperforms PURE~\cite{zhong2020frustratingly} on RE by 1.5\%/3.1\% and outperforms PFN~\cite{yan2021partition} on NER by 0.8\%/2.4\% on ACE05/SciERC respectively.
We also perform detailed analysis experiments to verify whether our framework can really achieve the mutual benefits of NER and RE. We observe that maintaining sequential inforamtion flow is especially crucial to the performance of both NER and RE. 
Moreover, compared with PFN~\cite{yan2021partition} which also consider implicit interaction, our method achieves remarkable improvement on NER, particularly for out-of-triple entity extraction. This also confirms that our sequential information propagation with the unified task form can eliminate the negative impacts RE imposes on NER. The above observations prove the superiority of our framework in the effective interaction of NER and RE.

%% file: paper/Related_Work.tex
Entity-relation extraction consists of two sub-tasks: named entity recognition (NER) and relation extraction (RE).
Traditionally, early pipeline methods \cite{sang2003introduction,florian2004statistical,chan2011exploiting} address it in two separate steps where NER and RE models are trained separately, which ignores the interaction between NER and RE. Recently, \citet{zhong2020frustratingly} proposed to fuse entity information to the RE model, realizing the sequential interaction between NER and RE. Their result reflects the direct benefits of NER to RE, which is reasonable because NER reduces the learning burden of the RE model. However, their method does not consider the potential benefits that RE may contribute to NER. 

Another branch of works propose cascade methods \cite{yu2019joint,wei2019novel}. These methods decompose relation triple extraction into subject entity extraction and object entity extraction, both formulated as span extraction. This leads to a unified label space. However, these works cannot deal with out-of-triple entity extraction. Our framework is partially motivated by this task setting, but we additionally incorporate an entity extraction model to enable the extraction of out-of-triple entities.

Both pipeline methods and decomposition-based methods suffer from error propagation and exposure bias problems. To address this, recent works seek to jointly solve the extraction of entities and relations. Some of them \cite{miwa2014modeling,gupta2016table,zhang2017end,tran2019neural,wang2020two,wang2020tplinker,wang2021unire} cast NER and RE as a table filling problem. They attempt to capture the implicit interaction between NER and RE by sharing the same encoder and weights for simultaneous prediction of entities and relations. However, they do not consider the possible contradiction in the shared information between entity and relation extraction due to the different task forms, which may harm the performance of individual tasks. 
\citet{yan2021partition} alleviated this issue with
a partition filter encoder to model two-way interaction between NER and RE. The encoder partitions the shared representation into two task-specific representations for NER and RE along with one shared representation. However, the partition operation only applies to the two-task situation, which limits the design of more sub-tasks.
In contrast, we propose a selection recurrent network which can handle the case of more than two sub-tasks and work well for our proposed task formulation.

In our work, we successfully combine the advantages of both implicit joint interaction and sequential interaction while alleviating their weaknesses.

%% file: paper/Method.tex
\section{Methodology}
\subsection{Problem Formulation}

Given a sentence $X$ consisting of $n$ tokens $x_1, x_2, . . . , x_n$, the first aim is named entity recognition (NER), to extract all the entities from $X$. 
An entity is denoted as $e=\langle x_i, x_j, \varepsilon_k \rangle$ where $x_i$ and $x_j$ are the start and end token of an entity and $\varepsilon_k\in \mathcal{E}$ denotes the entity type of data $e$. 
The second objective is relation extraction (RE), to identify relations between the recognized entities. We use $t=\langle e_i, e_j, r_k \rangle$ 
to represent a relation triple, which means entities $e_i$ and $e_j$ 
have the relation of relation type $r_k\in\mathcal{R}$. 
We regard $e_i$ as the subject and $e_j$ as the object in a relation triple.

Because some relation triples share the same entity as subjects, we model relations as functions $f(s_i, r_k) = \{o_1, o_2, ...\}$ that map subjects $s$ to objects $o$ given a specific relation type $r_k$, instead of performing relation classification directly. 
Therefore, our framework reformulates NER and RE as a multi-turn span extraction problem, including the following turns: entity extraction (EE), subject extraction (SE) and subject-oriented object extraction (SOE). We solve these three sub-tasks in a cascading manner. After obtaining subjects via SE and their corresponding objects through SOE, we transform them into relation triples $t$.

\subsection{Overview of the Framework}

We propose a unified cascade entity-relation framework which eliminates the gap between different task forms by using unified span extraction. It consists of three modules for EE, SE, and SOE. In all the modules, we first employ the same pre-trained BERT~\cite{devlin2018bert} encoder to obtain the contextual representations of the tokens in the sentence. Then we propose an SRN (Selection Recurrent Network) encoder which models implicit interaction and learns how to separate task-specific independent information and inter-task shared information in the token representations. We hence feed these task-specific representations from SRN to three span extraction models to obtain entity spans, subject spans along with corresponding object spans, respectively.

To maintain the sequential information flow between the sub-tasks, we also design two strategies for fusing the output information predicted by precedent span extraction modules into subsequent modules. To be more specific, when performing subject extraction, we make the module aware of the entity information extracted in EE. During SOE, the module is fed with both the entity information and subject information. We describe the details of sequential information propagation strategies in Section \ref{sec:strategy}. The overview of our framework with two strategies are illustrated in Figure \ref{fig:strategy}.

\subsection{Selection Recurrent Network}

\begin{figure}[t]
    \centering
    \includegraphics[width=0.92\linewidth]{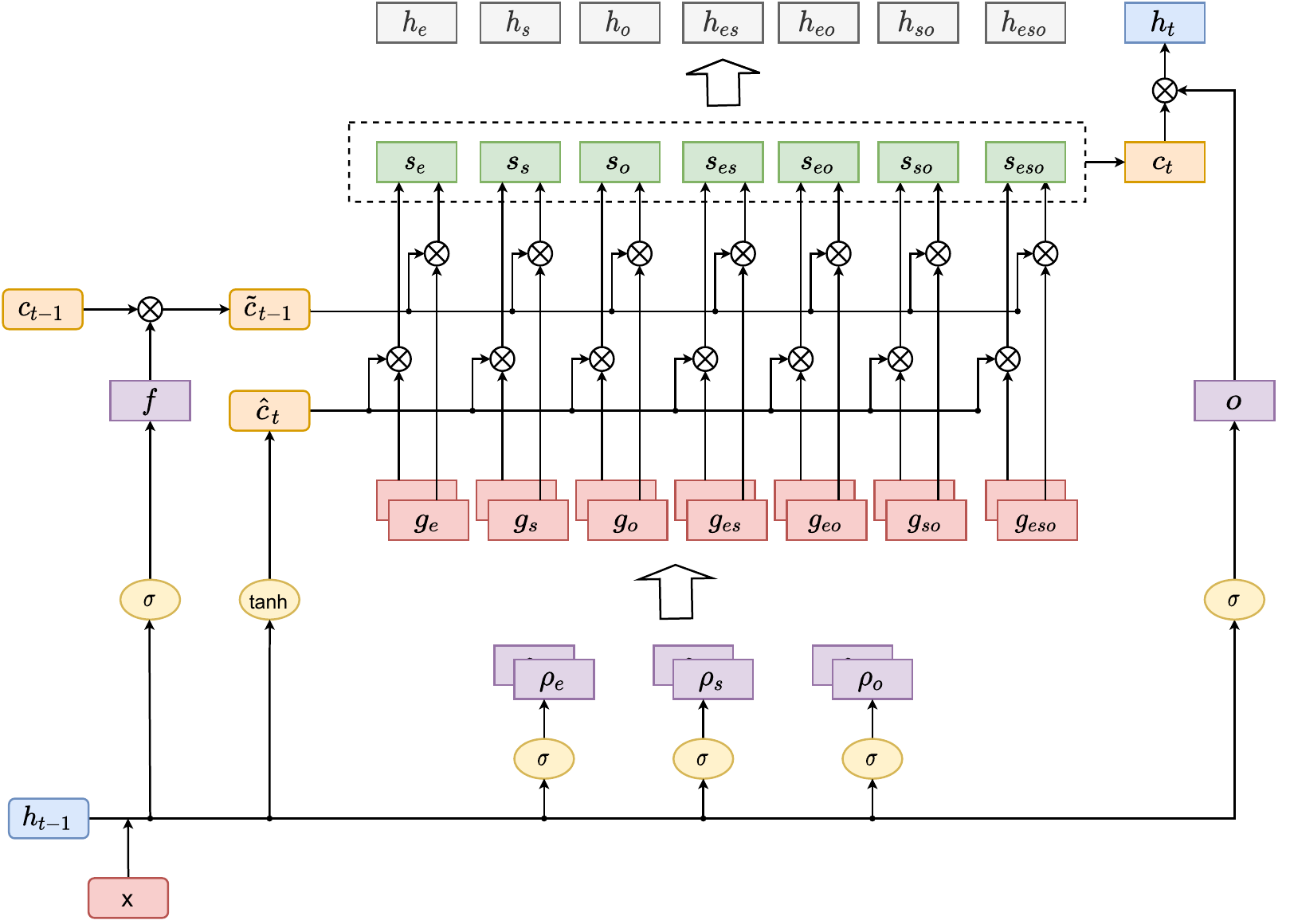}
    \caption{\label{fig:srn} The architecture of SRN. It is an LSTM-like network that encodes sequence information with cell states and hidden states which store intermediate information. At each time step, we perform selection operation to select task-relevant information from the intermediate information and calculate task-specific and task-shared representations as the inputs of task-specific modules. }
\end{figure}

We first introduce our proposed selection recurrent network (SRN) which is essential to learning the implicit interaction among multi-tasks. SRN is a novel RNN-based feature encoder. As shown in figure \ref{fig:srn}, this network is designed to jointly extract task-specific independent information and shared information from token representations. Similar to the standard LSTM, for each time step $t$, in addition to a hidden state $h_t$, we use an additional cell state $c_t$ to store intermediate memories.
We use a forget gate $f_t$ to control the forgetting operation for history cell state $c_{t-1}$, and use an output gate $o_t$ to generate hidden state $h_t$ from the current cell state $c_t$. We calculate the current state $\hat{c}$ and filtered history state $\tilde{c}_{t-1}$ with two gates as follows:

\begin{equation}
    \left[\begin{array}{l}
    f_{t} \\
    o_{t} \\
    \hat{c}_{t}
    \end{array}\right]=\left[\begin{array}{c}
    \sigma \\
    \sigma \\
    \tanh
    \end{array}\right] W_1 \cdot\left[h_{t-1}, x_{t}\right]
    \label{general_gates}
\end{equation}

\begin{equation}
    \tilde{c}_{t-1} = f_{t} \odot c_{t-1},
    \label{history}
\end{equation}

where $x_t$ is the input token representation for the current time step and $\sigma$ indicates sigmoid function. 
After obtaining $\tilde{c}_{t-1}$ and $\hat{c}$, we perform selection operations to generate three different filtered memories which store intra-task information and four different shared memories that store inter-task information.

Firstly, we calculate three selection master gates $\rho_{e,\hat{c}}$, $\rho_{s,\hat{c}}$, $\rho_{o,\hat{c}}$ for three sub-tasks. Master gates will select task-related neurons from state representations. An element of these three gates means whether the corresponding neuron belongs to the specific task. Similarly to computing forget gates and output gates, we generate the three master gates from history hidden state $h_{t-1}$ and input $x_t$.

\begin{equation}
    \left[\begin{array}{l}
    \rho_{e,\hat{c}} \\
    \rho_{s,\hat{c}} \\
    \rho_{o,\hat{c}} 
    \end{array}\right]=\left[\begin{array}{c}
    \sigma \\
    \sigma \\
    \sigma 
    \end{array}\right] W_2 \cdot\left[h_{t-1}, x_{t}\right]
    \label{eq:rho}
\end{equation}

Then, based on the master gates, we calculate four shared gates $g_{es, \hat{c}}$, $g_{eo, \hat{c}}$, $g_{so, \hat{c}}$, $g_{eso, \hat{c}}$ and three independent task gates $g_{e,\hat{c}}$, $g_{s,\hat{c}}$, $g_{o,\hat{c}}$. For example, $g_{eo, \hat{c}}$ represents shared gate for entity extraction and object extraction, while $g_{eso, \hat{c}}$ represents shared gate for all tasks
Each element in the four shared gates indicates whether the corresponding neuron is shared by multiple tasks, while each element of these independent gates means whether the corresponding neuron \textbf{only} belongs to the specific task.

\begin{equation}
    \begin{aligned}
        g_{es, \hat{c}} &=\rho_{e,\hat{c}} \odot \rho_{s,\hat{c}} \\
        g_{eo, \hat{c}} &=\rho_{e,\hat{c}} \odot \rho_{o,\hat{c}}\\
        g_{so, \hat{c}} &=\rho_{s,\hat{c}} \odot \rho_{o,\hat{c}}\\
        g_{eso, \hat{c}} &=\rho_{e,\hat{c}} \odot \rho_{s,\hat{c}} \odot \rho_{o,\hat{c}}
    \end{aligned}
    \label{eq:g_multi}
\end{equation}

\begin{equation}
    \begin{aligned}
        g_{e,\hat{c}} &= \rho_{e,\hat{c}} - g_{es, \hat{c}} - g_{eo, \hat{c}} + g_{eso, \hat{c}} \\
        g_{s,\hat{c}} &= \rho_{s\hat{c}} - g_{es, \hat{c}} - g_{so, \hat{c}} + g_{eso,\hat{c}} \\
        g_{o,\hat{c}} &= \rho_{o,\hat{c}} - g_{eo, \hat{c}} - g_{so, \hat{c}} + g_{eso, \hat{c}} 
    \end{aligned}
    \label{eq:g_single}
\end{equation}

As illustrated in Figure \ref{fig:srn}, we also compute the master gates, shared gates and independent task gates for filtered history state $\tilde{c}_{t-1}$ in exactly the same procedure as Equation \eqref{eq:rho} \eqref{eq:g_multi} \eqref{eq:g_single}. Hence we obtain $g_{e,\tilde{c}_{t-1}}$, $g_{o,\tilde{c}_{t-1}}$, $g_{s,\tilde{c}_{t-1}}$, $g_{es,\tilde{c}_{t-1}}$, $g_{eo,\tilde{c}_{t-1}}$, $g_{so,\tilde{c}_{t-1}}$,  $g_{eso,\tilde{c}_{t-1}}$.
\vspace{0.05mm}

Then, we compute the independent memories and shared memories to store the task-specific and shared information in the token representation to achieve flexible interaction among three sub-tasks. 

\begin{equation}
    \begin{aligned}
        s_e &=  g_{e,\tilde{c}_{t-1}} \odot \tilde{c}_{t-1} + g_{e,\hat{c}} \odot \hat{c} \\
        s_o &=  g_{o,\tilde{c}_{t-1}} \odot \tilde{c}_{t-1} + g_{o,\hat{c}} \odot \hat{c} \\
        s_s &=  g_{s,\tilde{c}_{t-1}} \odot \tilde{c}_{t-1} + g_{s,\hat{c}} \odot \hat{c}
    \end{aligned}
\end{equation}

\begin{equation}
    \begin{aligned}
        s_{es} &=  g_{es,\tilde{c}_{t-1}} \odot \tilde{c}_{t-1} + g_{es,\hat{c}} \odot \hat{c} \\
        s_{eo} &=  g_{eo,\tilde{c}_{t-1}} \odot \tilde{c}_{t-1} + g_{eo,\hat{c}} \odot \hat{c} \\
        s_{so} &=  g_{so,\tilde{c}_{t-1}} \odot \tilde{c}_{t-1} + g_{so,\hat{c}} \odot \hat{c} \\
        s_{eso} &=  g_{eso,\tilde{c}_{t-1}} \odot \tilde{c}_{t-1} + g_{eso,\hat{c}} \odot \hat{c}
    \end{aligned}
\end{equation}

Information stored in independent memories is inaccessible for other tasks. Instead, information stored in shared memories can be leveraged by different tasks. For example, $s_{es}$ stores the shared information between entity extraction and subject extraction. $s_{eso}$ stores the shared information of all three tasks.

Combining filterer memories and shared memories, we can update cell state $c_t$ and hidden state $h_t$ for next time step:
\vspace{-1mm}
\begin{equation}
        c_t = Linear([s_e;s_o;s_s;s_{es};s_{eo};s_{so};s_{eso}])
        \label{c_t}
\end{equation}
\vspace{-4mm}
\begin{equation}
        h_t = o_t \odot tanh(c_t)
        \label{h_t}
\end{equation}

Meanwhile, we generate task-specific independent token representations $h_e$, $h_s$, $h_o$ and shared token representations $h_{es}$, $h_{eo}$, $h_{so}$ and $h_{eso}$ from the corresponding memories:

\begin{equation}
    \left[\begin{array}{l}
    h_e \\
    h_s \\
    h_o \\
    h_{es}\\
    h_{eo}\\
    h_{so}\\
    h_{eso}
    \end{array}\right]=tanh(\left[\begin{array}{c}
    s_e \\
    s_s \\
    s_o \\
    s_{es}\\
    s_{eo}\\
    s_{so}\\
    s_{eso}
    \end{array}\right] )
\end{equation}

\subsection{Sequential Information Propagation}
\label{sec:strategy}

\begin{figure*}[t]
    \centering
    \subfigure[Early fusion]{
        \begin{minipage}[t]{0.45\linewidth}
            \centering
            \includegraphics[width=0.92\linewidth]{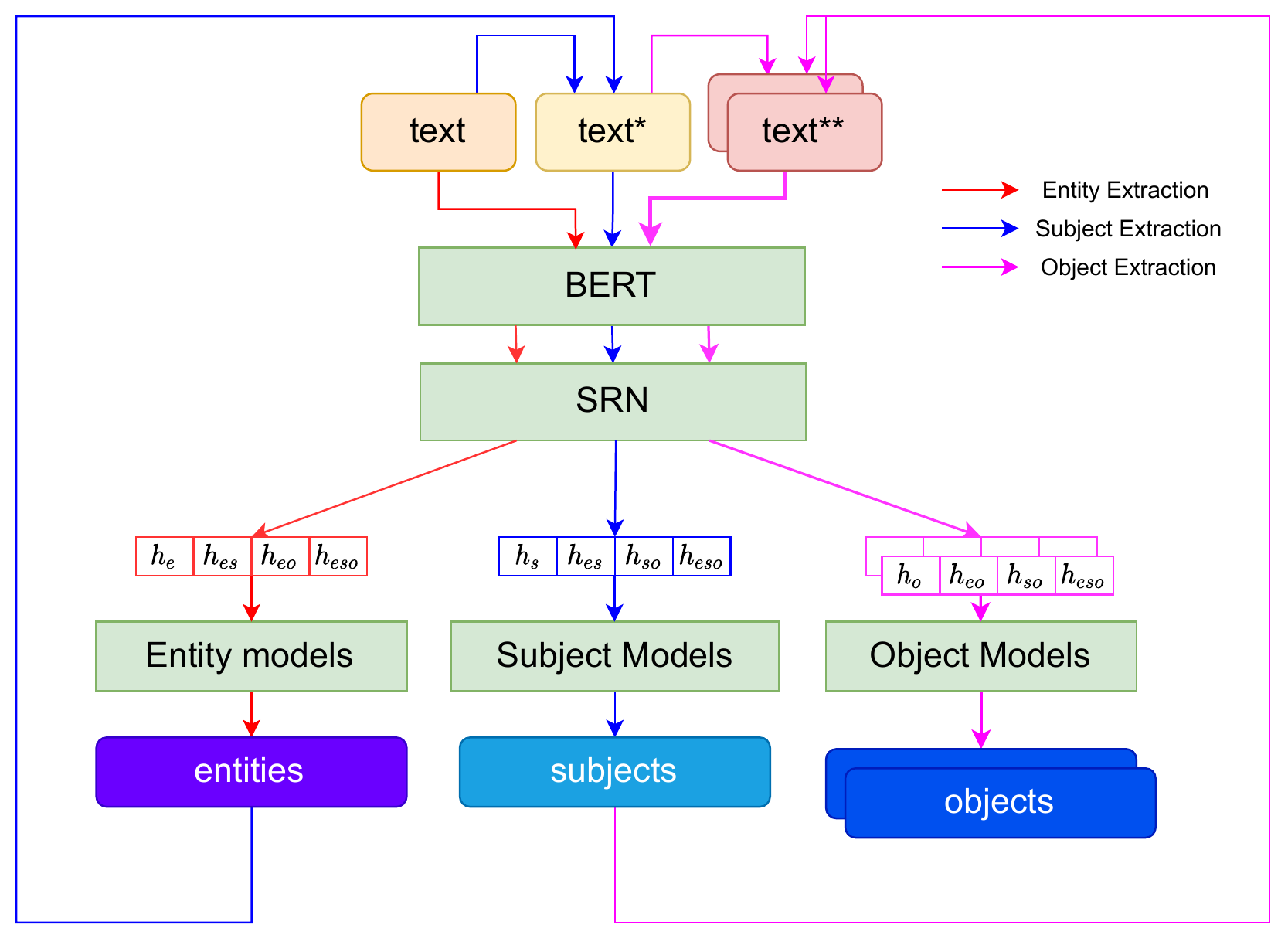}
        \end{minipage}
        \label{strategy_1}
    }
    \subfigure[Late fusion]{
        \begin{minipage}[t]{0.45\linewidth}
            \centering
            \includegraphics[width=0.92\linewidth]{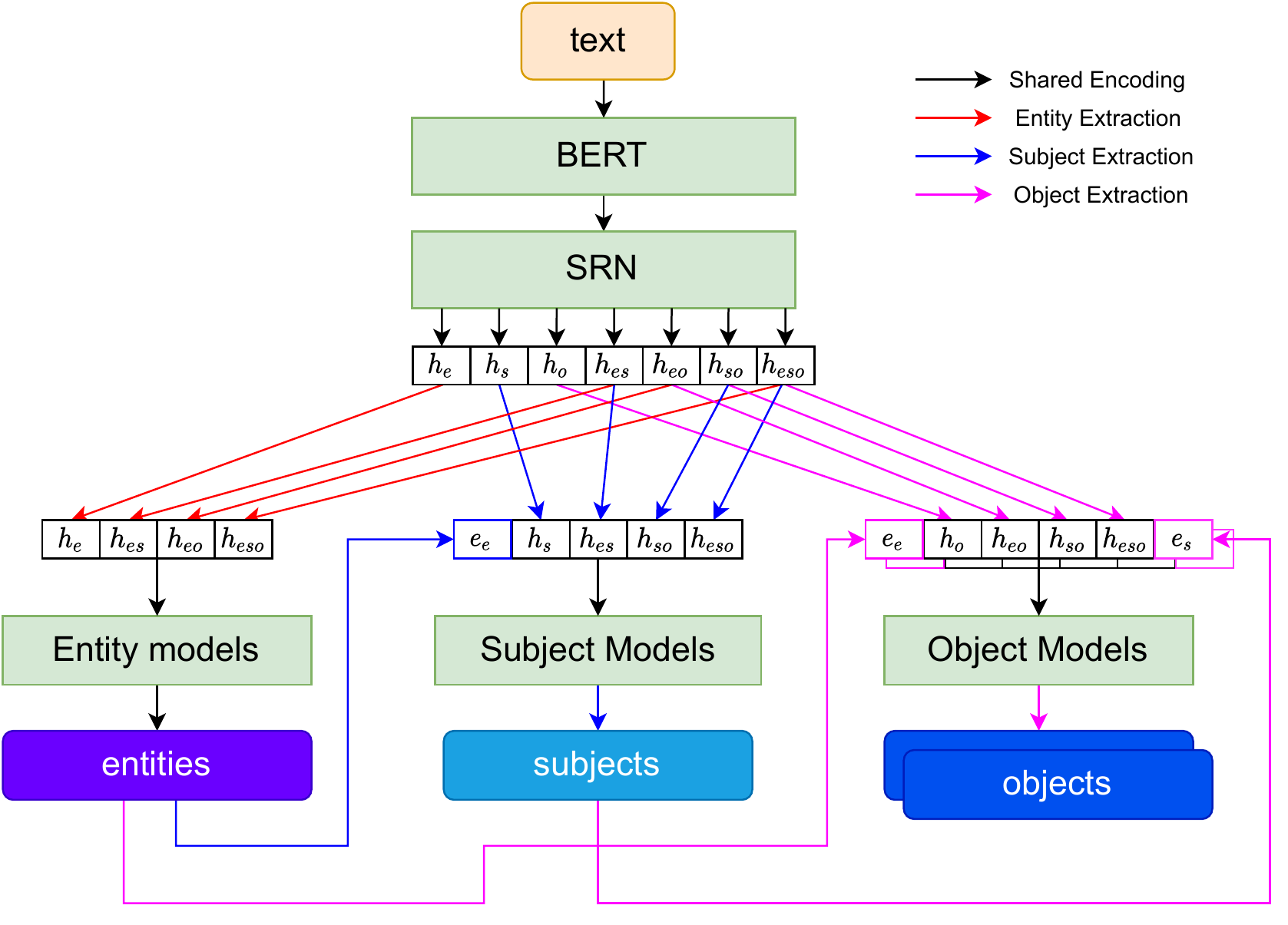}
        \end{minipage}
        \label{strategy_2}
    }
    \caption{\label{fig:strategy} An overview of our framework with early fusion and late fusion strategies. Different color indicators represent different processes in the sequential framework. 
    During inference, prior entity information and prior subject information for subject and object models is obtained from the output of entity models and subject models as shown in this figure. However, during training, we directly use ground-truth entity label information as the prior information as described in section \ref{section:train_and_inference}.
    In figure (a), ``text'' is the original input sentence, ``text*'' is text with inserted entity markers, ``text**'' is text with entity markers and subject markers. In figure (b), For entity, subject, object models, they share the same text representation for ``text'' encoded by BERT and SRN. 
    }
\end{figure*}


To maintain the sequential information flow between different tasks, we design two strategies to fuse the entity information in previous modules, namely, \textit{early fusion} and \textit{late fusion}. 

\subsubsection{Early Fusion}

Inspired by PURE~\cite{zhong2020frustratingly}, for the early fusion method, we insert pre-defined text markers into the input text right before and after the entities or given subject. We define text markers as \textit{[S:S]} and \textit{[S:E]} for the start and end of a subject separately, then insert them before and after the subject span to highlight the subject. By inserting pre-defined subject text markers, the model will not be confused by multiple potential subjects in the sentence. To highlight the entity with type $e_k$, we define text markers as \textit{[$e_k$\_S]} and \textit{[$e_k$\_E]} for the start and end of an entity span separately, then insert them before and after the entity span. With these text markers inserted in the input text, each of the span-extraction modules can easily understand the entity/subject information predicted by previous modules. As shown in Figure \ref{fig:strategy}~(a), the input sentences are updated with new text markers following the sequential prediction of the three task modules. For each module, only the task-specific token representation is obtained from the SRN encoder. For example, for entity extraction, token $x_i$ has one independent representation $h_{e,i}$ and three shared representation $h_{es, i}$, $h_{eo, i}$, $h_{eso,i}$.
After concatenating these representation, we get the final task-specific representation $\hat{h}_{e,i}$. In the same way, we get $\hat{h}_{s,i}$, $\hat{h}_{o,i}$. We then perform span extraction on these representations to obtain entity spans, subject spans, and object spans, respectively. 

\subsubsection{Late Fusion}

The weakness of the early fusion method is that
we need to encode the sentence every time for each turn of subject extraction and object extraction because each time the input sentence is modified by inserting different text markers. To reduce the time expense, we propose an alternative late fusion strategy. The key idea is that we would like to reuse the outputs of the encoder to save the repeating labor of text embedding and encoding. 
As shown in Figure \ref{fig:strategy}~(b), towards this end, we assign each token a subject tag to denote whether this token belongs to a subject span and meanwhile assign an entity tag to denote whether this token belongs to a certain type of entity or does not belong to any entity span.
We construct the entity-type embedding table as $\textbf{T}^{e}\in \mathbb{R}^{(\mathcal{E}+1)\times d^{t}}$, where $d^{t}$ is the dimension of the entity type embedding.
We construct the subject embedding table as $\textbf{T}^{s}\in \mathbb{R}^{2\times d^{t}}$. 
Next, given the information of entity and subject prediction, we can look up the two embedding tables to obtain corresponding entity embedding $e_{e,i}$ and subject embedding $e_{s,i}$ for each token $x_i$.
We concatenate the task-specific representations from the SRN encoder with the entity embeddings
as the input of the subject extraction module. For object extraction, we concatenate SRN outputs with both entity embeddings and subject embeddings.

\begin{equation}
    \begin{aligned}
        \hat{h}_{e, i} &= [h_{e, i}, h_{es, i}, h_{eo, i}, h_{eso,i}]
        \\
        \hat{h}_{s, i} &= [e_{e, i}, h_{s, i}, h_{es, i}, h_{so, i}, h_{eso,i}]
        \\
        \hat{h}_{o, i} &= [e_{e, i}, h_{o, i}, h_{eo, i}, h_{os, i}, h_{eso,i}, e_{s, i}]
    \end{aligned}
\end{equation}

Finally, we will perform span extraction given these input representations to obtain entity/subject/object spans.

\subsection{Span Extraction Model}

Entity span extraction, subject span extraction, and object span extraction are all performed by using a span extraction model with a unified architecture inspired by the span selection in machine reading comprehension~\cite{li2020unified}. 
Specifically, we first adopt two kinds of identical binary sequence classifiers to detect the start and end position of entities/subjects/objects respectively by assigning each token a binary tag (0/1) that indicates whether the current token corresponds to a start or end position of a target span. Then, we use a token-pair classifier to match the detected start and end positions as a span.

%% file: paper/Experiments.tex

\subsection{Experiment Settings}

\subsubsection{Datasets}
We evaluate our methods on two standard entity-relation extraction benchmarks: ACE05~\cite{walker2006ace} and SciERC~\cite{luan2018multi}, ACE05 is collected from various domains, including newswire and online forums. SciERC is collected from AI paper abstracts and contains annotations of scientific entities and their relations. 

\subsubsection{Evaluation metrics}

We evaluate our models under strict evaluation protocol following previous work~\cite{zhong2020frustratingly, yan2021partition} and use micro F1 measure as the evaluation metric. For NER, a predicted entity is considered as a correct prediction if its span boundaries and the predicted entity type are both correct. For RE, a predicted relation is considered as correct only if the boundaries and entity type of both subject and object spans are correct and the predicted relation type is correct.

\subsection{Main Results}

\begin{table*}[t]
  \centering
  \resizebox{1.5\columnwidth}{!}{
    \begin{tabular}{l|c|c|c|c|c}
    \toprule [0.7 pt]
    \multirow{2}[0]{*}{Method} & \multirow{2}[0]{*}{Embedder}& \multicolumn{2}{c|}{ACE05} & \multicolumn{2}{c}{SciERC} \\
    \cline{3-6}
          & & NER & RE & NER & RE \\
    \midrule[0.7 pt]
    Structured Perceptron \cite{li2014incremental} & - & 80.8  & 49.5  & -     & - \\
    \hline
    SPTree \cite{miwa2016end} &Word2vec & 83.4  & 55.6  & -     & - \\
    \hline
    Multi-turn QA \cite{li2019entity} & Bert-large & 84.8  & 60.2  & -     & - \\
    \hline
    \multirow{2}[0]{*}{SPE\cite{wang2020pre}} & SPE & 87.2  & 63.2  & -    & - \\
     & SciBERT & - & - & 68.0    & 34.6\\
     \hline
    Table-Sequence \cite{wang2020two} & ALBERT & 89.5  & 64.3  & -     & - \\
    \hline
    \multirow{2}[0]{*}{PURE \cite{zhong2020frustratingly}} & ALBERT & 89.7  & 65.6  & -  & - \\
     & SciBERT & -  & -  & 66.6  & 35.6 \\
     \hline
    \multirow{2}[0]{*}{PFN \cite{yan2021partition}} &ALBERT & 89.0    & 66.8  & -  & - \\
    &SciBERT & -    & -  & 66.8  & 38.4 \\
    \midrule[0.7 pt]
    \multirow{2}[0]{*}{Cascade-SRN (Late fusion)}  &ALBERT & 89.4  & 65.9  & -  & - \\
    & SciBERT & -  & -  & 68.6  & 36.7 \\
    \hline
    \multirow{2}[0]{*}{Cascade-SRN (Early fusion)}   &ALBERT & \textbf{89.8}  & \textbf{67.1}  & -  & -\\
    &SciBERT & -  & -  & \textbf{69.2}  & \textbf{38.7} \\
    \bottomrule [0.7pt]
    \end{tabular}%
    }
    \caption{Main results (\%): F1 scores on test splits of ACE05 and SciERC. Results of PURE are reported in single-sentence setting for fair comparison.}
    \label{tab:main_result}
\end{table*}%

\subsubsection{Results and analysis}

Table \ref{tab:main_result} shows the comparison of our methods with previous results. We report the F1 scores in single-sentence setting. 
As it is shown, the proposed method Cascade-SRN using the late fusion strategy shows competitive performance for NER and RE compared to previous methods. While the proposed method Cascade-SRN using the early fusion strategy outperforms all the baselines. We state that early fusion strategy achieves a better F1 score because it makes bert-based embedder and SRN encoder aware of the prior information so that it has a larger learning space. 
Because the method with early fusion achieves higher performance, we regard it as our default model.
Compared with the previous best pipeline method PURE \cite{zhong2020frustratingly}, our model outperforms by 3.1\% on SciERC for RE and 1.5\% on ACE05, while obtaining 2.6\% boost of NER F1 on SciERC but only 0.1\% NER F1 on ACE05. This shows that compared with the pipeline model with a separate encoder, our method mainly brings the improvement of RE performance.
Compared with the previous best joint method PFN \cite{yan2021partition}, our model outperforms by 0.8\% on ACE05 and 2.6\% on SciERC for NER, while we obtain 0.3\% RE F1 on ACE05 and SciERC.

Combining these observations, we find that our method achieve more balanced performance on NER and RE compared with previous pipeline and joint methods,
Because previous pipeline methods do not model the mutual interaction between NER and RE, they do not leverage NER information to help RE prediction. Meanwhile, the previous joint method cannot deal with the dynamic between in-triple and out-of triple entity prediction, so importing RE information sometimes even hurts the final NER score. Our methods averts these two drawbacks and successfully makes RE information beneficial to entity prediction.

\subsection{Ablation Study}

\begin{table}[t]
\setlength{\tabcolsep}{1.4mm}
\small
  \centering
    \begin{tabular}{c|c|cc}
    \hline
    Ablation & Settings & NER & RE\\
    \hline
    \multirow{3}[0]{*}{Encoder} & SRN & \textbf{69.2}  & \textbf{38.7} \\
        & BiSRN & 69.1 & 36.8\\
        & BiLSTM & 68.2 & 36.5 \\
    \hline
    \multirow{2}[0]{*}{Text Marker}
        & Entity+Subject & \textbf{69.2} & \textbf{38.7} \\
        & Subject & 66.2 & 35.6 \\
    \hline
    \multirow{2}[0]{*}{\makecell[c]{Training \\strategy}}
        & golden entity & \textbf{69.2} & \textbf{38.7} \\
        & golden subject & 68.1 & 33.7 \\
    \hline
    \multirow{2}[0]{*}{\makecell[c]{Joint Vs. \\Independent}}
        & Joint & \textbf{69.2} & 38.7 \\
        & Independent & 67.5 & - \\
    \hline
    \end{tabular}%
\caption{Ablation study results.}
\label{tab:ablation}%
\end{table}%

\begin{table*}[t]
\setlength{\tabcolsep}{1.2mm}
\small
\centering
\caption{Entity prediction results on different entity groups. Entities are split into two groups: In-triple and Out-of-triple based on whether they appear in relation triples or not.}
\label{tab:entity}%
    \begin{tabular}{c|ccc|ccc|ccc|ccc}
    \hline
    \multirow{3}[0]{*}{Model} & \multicolumn{6}{c|}{SciERC} &\multicolumn{6}{c}{ACE05}\\ 
    \cline{2-13}
    &\multicolumn{3}{c|}{In-triple}   &\multicolumn{3}{c|}{Out-of-triple}               
    & \multicolumn{3}{c|}{In-triple} & \multicolumn{3}{c}{Out-of-triple}\\
        \cline{2-13}
          & P     & R     & F1    & P     & R     & F1    & P     & R     & F1    & P     & R     & F1 \\
    \hline
    PFN \cite{yan2021partition}  & 78.0 & 71.1 & 74.4 & 38.9 & 61.7 & 47.8  & \textbf{95.9}  & 92.1  & \textbf{94.0}  & 85.8  & 86.9  & 86.3 \\
    \hline
    Cascade-SRN (late fusion) & \textbf{82.1} & \textbf{71.7} & \textbf{76.5} & 39.5 & \textbf{66.2} & 49.5 & 95.7 & 91.9 & 93.8  & 86.2  & 87.8  & 87.0 \\
    Cascade-SRN (early fusion) & 82.0 & 70.3 & 75.7 & \textbf{44.5} & 61.5 & \textbf{51.6} & 95.4 & \textbf{92.3} & 93.8  & \textbf{86.5}  & \textbf{88.1}  & \textbf{87.3} \\
    \hline
    \end{tabular}%
\end{table*}%

In this section, we perform ablation studies on the SciERC dataset to analyze the effects of our method with respect to different aspects: Encoder, Text Marker, Training Strategy, and Joint Vs. Independent. Table \ref{tab:ablation} reports the ablation study results.

\subsubsection{Encoder}
We replace our SRN encoder with Bidirectional SRN or BiLSTM. Our SRN encoder is a unidirectional encoder which only in the forward direction. Bidirectional SRN includes two SRN encoders, one in the forward direction and the other one in the backward direction. We observe replacing unidirectional SRN with Bidirectional SRN cannot bring performance boost.
When we replace SRN with BiLSTM, the experiment results show an obvious drop in NER and RE performance. This proves the effectiveness of the SRN model and the importance of filtering irrelevant information in the shared representation.

\subsubsection{Text Marker}
In our early fusion strategy for sequential information propagation, we insert text markers into the input sentence of each span extraction module. Here we aim to validate the effects of entity markers which maintain the sequential information flow between NER and RE. We compare two settings: Entity+Subject Marker and Subject Marker. In the former setting, we insert entity type markers into the input sentences of the subject module and for the input of the object module, both entity markers and subject markers are inserted. This is the default setting in our method. As for the latter setting, we feed the original sentence without entity markers into the subject module and only insert subject markers into the input sentence of the object module. This setting removes the information propagation from NER to RE. We observe that the first setting outperforms the second one by 3.0\% on NER F1 and 3.1\% on RE F1, proving the effectiveness of our fusion strategy and the importance of sequential information propagation between NER and RE. However, it is interesting that without this one-way information flow, the NER performance also decreases a lot. The possible reason is that without entity information, the RE (subject and object) modules have to do the extra work of entity extraction implicitly by itself, which can disturb the training of the NER module since the gradients are back-propagated through a shared encoder. Note that we cannot remove subject markers, because our object module is subject-oriented. 

\subsubsection{Training Strategy}
When training object extraction modules, our default setting is to regard all ground-truth entities in the training dataset as candidate subjects, as described in Section \ref{section:train_and_inference}. Here we test the performance of using only ground-truth subjects. Results in Table \ref{tab:ablation} shows that only using golden subjects degrades the RE F1 score by 5\%. This demonstrates that using all gold entities can alleviate the exposure bias problem of the object module by enabling the model to learn how to identify incorrect subject spans.

\subsubsection{Joint Vs Independent}
To explore whether RE helps NER, we compare the joint model with the independent NER model. Using the same model architecture, we test the performance of a single entity extraction module without jointly training subject extraction and object extraction. We observe that the joint model outperforms the single independent model on NER F1 score, proving the contribution of the RE information to NER.

\subsection{Analysis}

\subsubsection{Analysis on entity prediction}

To understand why our methods achieve better NER results than previous joint methods, especially PFN \cite{yan2021partition} which also models inter-task interaction, we follow \citet{yan2021partition} to design comparison experiments on NER by splitting entities into two groups: in-triple entities and out-of-triple entities. In-triple entities appear in relation triples in the dataset while out-of-triple have no relations with other entities. For ACE05, 64\% entities in the test split are out-of-triple while the rest are in-triple. For SciERC, 22\% entities in the test set are out-of-triple entities while the others are in-triple.

As shown in table \ref{tab:entity}, for out-of-triple entities, our model with the early fusion strategy outperforms PFN by 3.8\% on sciERC and 1.0\% on ACE05. As for in-triple results, our methods outperform PFN on SciERC and achieve close performance on ACE05. This demonstrates the superiority of our model in handling out-of-triple entities and the effectiveness of our multi-task interaction.

\subsubsection{Analysis on fusion strategies}

\begin{table}[t]
\small
  \centering
    \begin{tabular}{l|cc|c}
    \hline
    Strategy & NER & RE & Speed (sentence/s) \\
    \hline
    early fusion & \textbf{69.2} & \textbf{38.7} &  9.04\\
    late fusion & 68.6 & 36.7 &  \textbf{295.04}\\
    \hline
    \end{tabular}%
  \caption{Comparison of F1 performance (\%) and inference speed on SciERC test sets. The speed is measured on a single NVIDIA GeForce 2080Ti GPU with a batch size of 32.}
\label{tab:strategies}%
\end{table}%

As shown in table \ref{tab:strategies}, We evaluate the inference speed of the two sequential information propagation strategies in our methods. We employ inference with early fusion strategy and late fusion strategy on SciERC test set with batch size 32 and SciBERT as the pre-trained embedder. We observe that the model with a late fusion strategy obtains a $32.7\times$ speedup compared with early fusion. This is because the late fusion strategy re-uses the embedder and encoder results which saves a lot of computations. Meanwhile, the model with early fusion achieves better performance by sacrificing time efficiency.

%% file: paper/Appendix.tex
\appendix

\section{Span Extraction Model}

Entity span extraction, subject span extraction, and object span extraction are all performed by using a span extraction model with a unified architecture inspired by the span selection in machine reading comprehension~\cite{li2020unified}. 
Specifically, we first adopt two kinds of identical binary sequence classifiers to detect the start and end position of entities/subjects/objects respectively by assigning each token a binary tag (0/1) that indicates whether the current token corresponds to a start or end position of a target span. Then, we use a token-pair classifier to match the detected start and end positions as a span. 
Here, we take entity extraction as an example to explain the process of span extraction. 

\paragraph{Span Start/End Prediction}

\begin{figure}[t]
    \centering
    \includegraphics[width=0.92\linewidth]{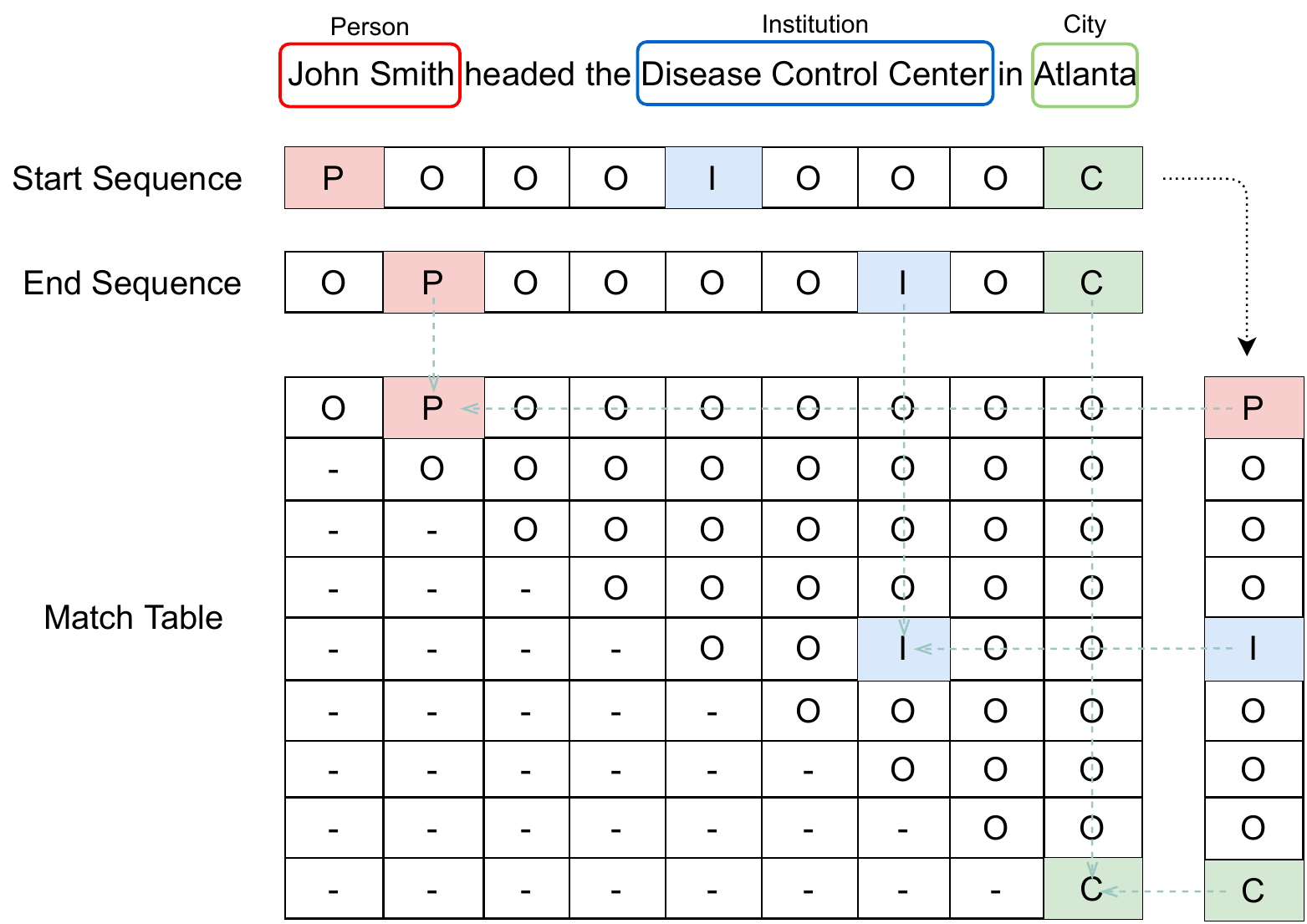}
    \caption{An example of span extraction output for entity extraction. We combine outputs of different entity types into one output space in this figure for ease of illustration. In the proposed span extraction model, two kinds of binary sequence classifiers are leveraged to predict the start and end sequence while a matching classifier is used to predict the start-end match table. \textit{P} is short for \textit{Person}. \textit{I} is short for \textit{Institution}. \textit{C} is short for \textit{City}. \textit{O} is short for \textit{Others}.}
    \label{fig:span}
\end{figure}

Given the task-specific representation $\hat{h}_e$, we use a token-level classifier composed of a linear layer and a sigmoid function, which calculates the probability of the labels of tokens for each entity type $e_k$:

\begin{equation}
    \begin{aligned}
    \hat{\textbf{y}}_{e_{st}}(x_i, e_k) &= \sigma(\textbf{W}_{st,k}\hat{\textbf{h}}_{e,i}+\textbf{b}_{st,k})
    \\
    \hat{\textbf{y}}_{e_{en}}(x_j, e_k) 
    &= \sigma(\textbf{W}_{en, k}\hat{\textbf{h}}_{e,j}+\textbf{b}_{en,k}),
    \end{aligned}
\end{equation}

where $\textbf{W}_{st,k}$, $\textbf{b}_{st,k}$, $\textbf{W}_{en,k}$ and $\textbf{b}_{en,k}$ are trainable parameters of the classifier, $\hat{\textbf{y}}_{e_{st}}(x_i, e_k)$ and $\hat{\textbf{y}}_{e_{en}}(x_j, e_k)$ represent the probability of classifying token $x_i$ and $x_j$ as the start and end token of an entity span of type $e_k$, respectively.

\paragraph{Start-End Matching}

One sentence may include multiple entities of the same entity type. Hence multiple start tokens and multiple end tokens may be predicted from the classifiers. We additional develop a matching model to match the start token with its end token.
The design of the matching model is similar to the table filling method \cite{yan2021partition} as shown in Figure \ref{fig:span}.
For each token pair $(x_i, x_j)$, we concatenate token-level entity features $\hat{\textbf{h}}_{e,i}$ and $\hat{\textbf{h}}_{e,j}$, then feed them into a fully-connected layer with sigmoid function to calculate the probability of token $x_i$ and $x_j$ being the start and end position of the same $e_k$-type entity.

\begin{equation}
    \hat{\textbf{y}}_{e_{m}}(x_i, x_j, e_k) 
    = \sigma(\textbf{W}_{m, k}\cdot [\hat{\textbf{h}}_{e,i}, \hat{\textbf{h}}_{e,j}]+\textbf{b}_{m,k})
\end{equation}

Subject and object extraction is the same as the above entity extraction model. Similarly, we obtain subject predictions $\hat{\textbf{y}}_{s_{st}}(x_i)$, $\hat{\textbf{y}}_{s_{en}}(x_j)$ and $\hat{\textbf{y}}_{s_{m}}(x_i, x_j)$ by subject model, obtain object results $\hat{\textbf{y}}_{o_{st}}(x_i, r_k, s_l)$, $\hat{\textbf{y}}_{o_{en}}(x_j, r_k, s_l)$ and $\hat{\textbf{y}}_{o_{m}}(x_i, x_j, r_k, s_l)$ for given relation type $r_k$ and given subject $s_l$ by object model.

\section{Training and Inference}
\label{section:train_and_inference}

During training process, the overall loss function $L$ is the sum of $L_e$, $L_s$ and $L_o$. The entity extraction module is trained via $L_e$ consisting of three parts: $L_{e,st}$, $L_{e,en}$, $L_{e, m}$, which represent the loss of entity start, entity end and start-end match respectively.
Similarly, $L_s$ trains the subject module and $L_o$ trains the object module.

\begin{equation}
    \begin{aligned}
    L_{e, st} &=-\sum_{i=1}^{n}\sum_{k=1}^{|\mathcal{E}|}BCELoss\\&(
    \textbf{y}_{e, st}(x_i,e_k),\hat{\textbf{y}}_{e,st}(x_i, e_k))
    \\
    L_{e, en} &=-\sum_{j=1}^{n}\sum_{k=1}^{|\mathcal{E}|}BCELoss\\&(
    \textbf{y}_{e, en}(x_j,e_k),\hat{\textbf{y}}_{e,en}(x_j, e_k))
    \\
    L_{e, m} &=-\sum_{i=1}^{n}\sum_{j=1}^{n}\sum_{k=1}^{|\mathcal{E}|}BCELoss\\&(
    \textbf{y}_{e, m}(x_i, x_j, e_k),\hat{\textbf{y}}_{e,m}(x_i, x_j, e_k)),
    \end{aligned}
\end{equation}

where $\textbf{y}_{e, st}(x_i,e_k)$ represents the binary label of whether token $x_i$ is the start of an $e_k$-type entity span, $ \textbf{y}_{e, sn}(x_j,e_k)$ represents the label of whether token $x_j$ is the end of an $e_k$-type entity span, and $ \textbf{y}_{e, m}(x_i, x_j, e_k)$ represents start-end matching label for word pair $(x_i, x_j)$. Similarly, we can compute the subject extraction loss $L_s$ and object extraction loss $L_o$ and the overall training objective is to be minimized the sum of the three span extraction losses:
\begin{equation}
\begin{aligned}
L_e &= \alpha_e L_{e, st} + \beta_e L_{e, en} + \gamma_e L_{e, m}
\\
L_s &= \alpha_s L_{s, st} + \beta_s L_{s, en} + \gamma_s L_{s, m}
\\
L_o &= \alpha_o L_{o, st} + \beta_o L_{o, en} + \gamma_o L_{o, m}
\\
L &= L_e + L_s + L_o,
\end{aligned}
\end{equation}
where $\alpha_e, \beta_e, \gamma_e, \alpha_s, \beta_s, \gamma_s, \alpha_o, \beta_o, \gamma_o$ are weighting hyper-parameters in the range 0-1 to balance the effects of start/end/matching losses of each span extraction module.

For training the subject module and object module, we only consider the ground-truth entity labels as the input. In order to mitigate the exposure bias problem when training the object module, we regard all ground-truth entities (including those are not subjects) in the training data as candidate subjects 
and perform object extraction for each of them. 

During inference, the entity extraction module first predicts the entities in the input sentence, then the predicted entity information is fused into the subject module to predict all candidate subjects. Finally, the object module extracts objects for these predicted subjects one by one, while being aware of the entity and subject information predicted by prior modules.

\section{Datasets}

\begin{table}[h]
\caption{Statistics of datasets. $\mathcal{E}$ and $\mathcal{R}$ are numbers of entity and relation types.}
 \label{tab:dataset}%
  \centering
    \begin{tabular}{cccccc}
    \hline
    \multirow{2}[0]{*}{Dataset} & \multicolumn{3}{c}{\#sentences} & \multirow{2}[0]{*}{|$\mathcal{E}$|} & \multirow{2}[0]{*}{|$\mathcal{R}$|} \\
          & Train & Dev & Test &       &  \\
    \hline
    ACE05 & 10051 & 2424  & 2050  & 7     & 6 \\
    SciERC & 1861  & 275   & 551   & 6     & 7 \\
    \hline
    \end{tabular}%
\end{table}%

We evaluate our methods on two standard entity-relation extraction benchmarks: ACE05~\cite{walker2006ace} and SciERC~\cite{luan2018multi}, ACE05 is collected from various domains, including newswire and online forums. SciERC is collected from AI paper abstracts and contains annotations of scientific entities and their relations. 
The statistics of the datasets are given in Table \ref{tab:dataset}. 

We do not evaluate on another common dataset ACE04 because it does provide official train/val/test splits and easily renders the evaluation results incomparable across different previous works. Moreover, the data distribution of ACE04 overlaps with ACE05.

\section{Implementation details}

We leveraged pre-trained language model BERT \cite{devlin2018bert} as the text embedder in our methods. Following previous work \cite{wang2020two, zhong2020frustratingly, yan2021partition}, the versions we use are albert-xxlarge-v1 for ACE05 and scibert-scivocab-uncased for SciERC. 
Training batch size and learning rate are set to 8 and 1e-5, respectively.
We set the hidden state dimension of the SRN encoder to 128.  
The model parameters were optimized by Adam optimizer \cite{kingma2014adam} for 100 epochs.
Also, to prevent gradient explosion, gradient clipping is applied during training.
The model checkpoint is validated with an interval of 100 steps and selected based on the best sum of NER F1 score and RE F1 score on the validation set.
All experiments are conducted on a single Tesla-P100 GPU or single NVIDIA GeForce 2080Ti GPU.

\section{Error Analysis of Relation Prediction}

\begin{figure}[t]
    \centering
    \includegraphics[width=0.92\linewidth]{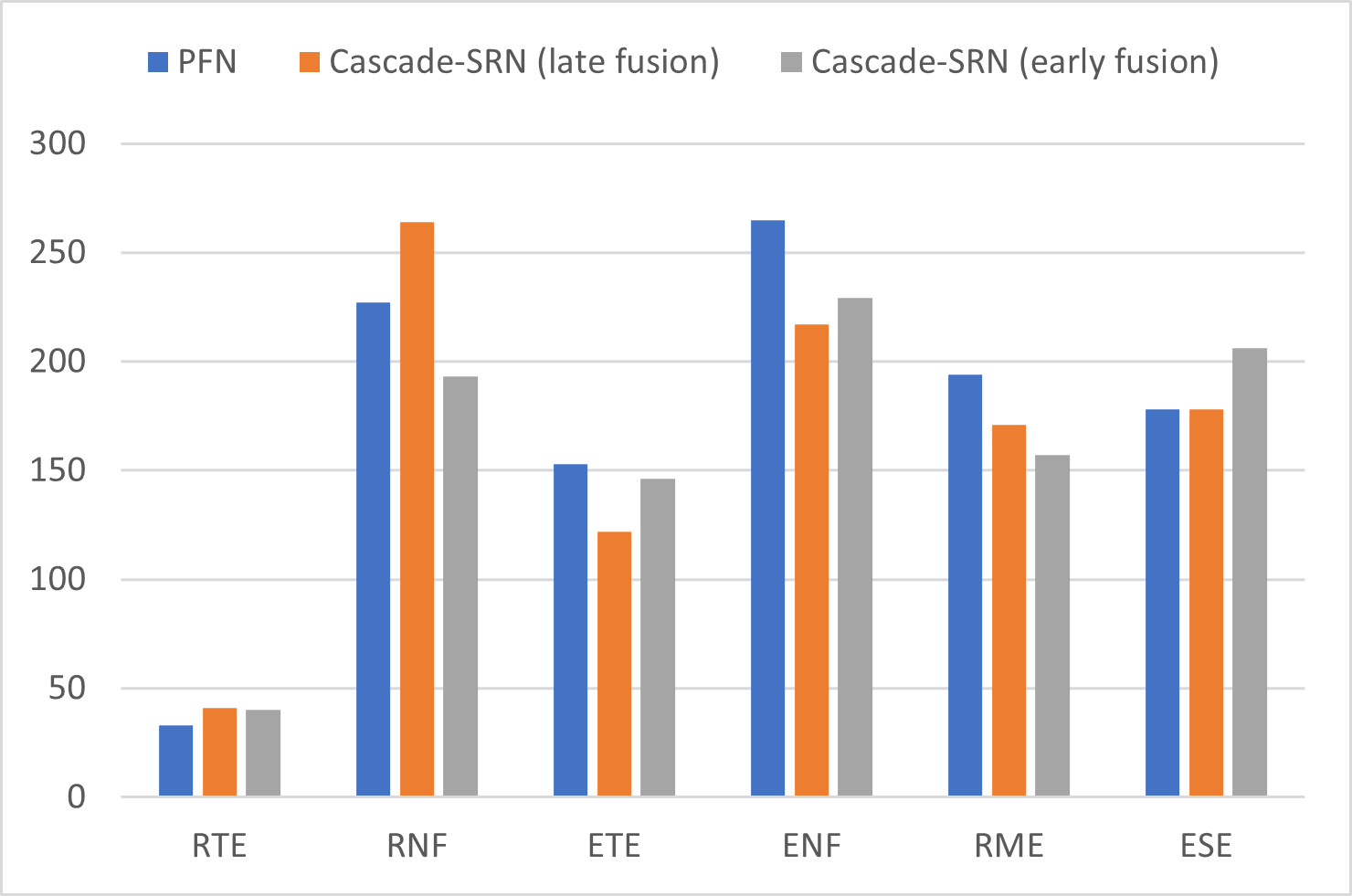}
    \caption{\label{fig:error}Distribution of six types of relation extraction errors on SciERC test set.}
\end{figure}

Our methods have similar RE performance to PFN \cite{yan2021partition}. We explore the specific difference by analyzing the error types for relation extraction.
We split these errors into six groups: (1) Relation type error (RTE) means only the relation type of predicted relation triple is wrong. (2) Relation not found (RNF) means the golden relation which is not included in predicted relation triples. (3) Entity type error (ETE) means the entity type of subject or object in predicted relation triple is wrong. (4) Entity not found (ENF) means the subject or object in golden relation is not extracted by NER model. (5) Relation match error (RME) means the subject and object of the predicted relation triple are correct entities but they do not have the predicted relation. (6) Entity span error (ESE) means the subject or object of predicted relation is a wrong entity span.

Figure \ref{fig:error} presents the distribution of the six errors on SciERC test set for PFN \cite{yan2021partition}, Cascade-SRN (late fusion) and Cascade-SRN (early fusion). Compared with PFN \cite{yan2021partition} and Cascade-SRN (late fusion), Cascade-SRN (early fusion) has fewer RNF and RME errors. These two error types are directly related to the effect of relation extraction, which proves the effectiveness of our framework and the early fusion strategy. However, early fusion strategy makes embedder and encoder aware of prior entity information for subject and object modules, so this makes our models more sensitive to entity prediction results, leading to more ESE errors.

\section{Selection Recurrent Network for Two Subtasks}

\begin{figure}[t]
    \centering
    \includegraphics[width=0.8\linewidth]{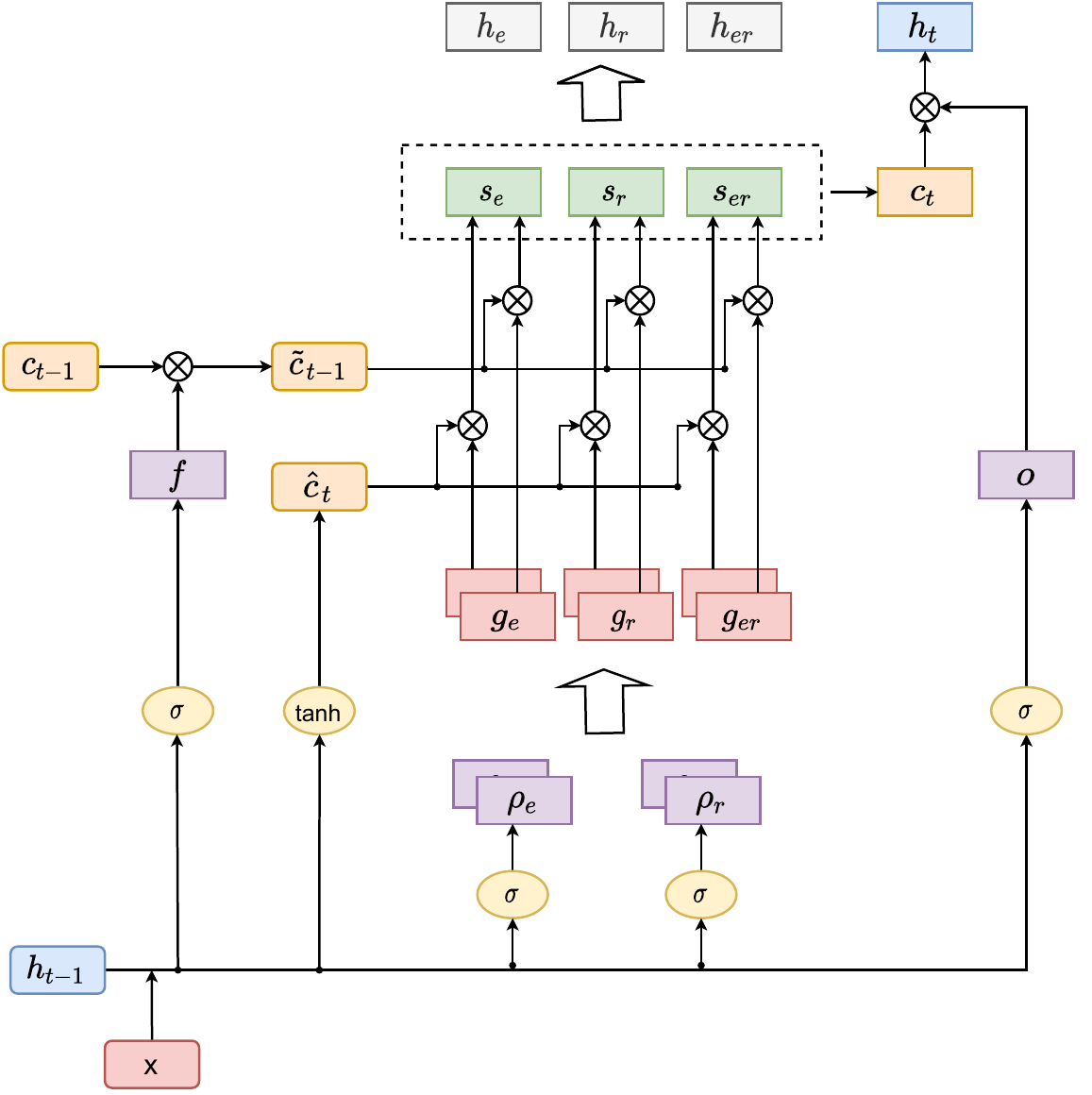}
    \caption{\label{fig:srn_2}Architecture of SRN for two subtasks.}
\end{figure}

In this section, we present the detail implementation of Selection Recurrent Network (SRN) for two subtasks (entity extraction and relation extraction). 
We calculate current state $\hat{c}$ and filtered history state $\tilde{c}_{t-1}$ with forget and output gates $f_{t}, o_{t}$ as Equations \ref{general_gates}, \ref{history}:

After obtaining $\tilde{c}_{t-1}$ and $\hat{c}$, we perform selecting operation to generate two different filtered memory which store intra-task information and one shared memory who store inter-task information.

Firstly, we calculate three master gates $\rho_{e,\hat{c}}$, $\rho_{r,\hat{c}}$, for two subtasks. Master gates will select task-related neuron from state representation. The element of these two gates means if corresponding neuron is belong to specific task. Like forget gate and output gate, we generate two master gates from history hidden state and input representation.

\begin{equation}
    \left[\begin{array}{l}
    \rho_{e,\hat{c}} \\
    \rho_{r,\hat{c}} 
    \end{array}\right]=\left[\begin{array}{c}
    \sigma \\
    \sigma 
    \end{array}\right] W_2 \cdot\left[h_{t-1}, x_{t}\right]
\end{equation}

Then, based on these two master gates, we calculate one shared gates $g_{er, \hat{c}}$, and two independent task gates $g_{e,\hat{c}}$, $g_{r,\hat{c}}$. 

\begin{equation}
    \begin{aligned}
        g_{er, \hat{c}} &=\rho_{e,\hat{c}} \odot \rho_{r,\hat{c}} \\
        g_{e,\hat{c}} &= \rho_{e,\hat{c}} - \rho_{er, \hat{c}}  \\
        g_{r,\hat{c}} &= \rho_{r\hat{c}} - \rho_{er, \hat{c}} 
    \end{aligned}
\end{equation}

To model interaction between two subtasks, information stored in independent memory is inaccessible for other tasks. Instead, information stored in shared memory can be leveraged by different tasks. 

\begin{equation}
    \begin{aligned}
        s_e &=  g_{e,\hat{c}} \odot \tilde{c}_{t-1} + g_{e,\hat{c}} \odot \hat{c} \\
        s_r &=  g_{r,\hat{c}} \odot \tilde{c}_{t-1} + g_{r,\hat{c}} \odot \hat{c} \\
        s_{er} &=  g_{er,\tilde{c}_{t-1}} \odot \tilde{c}_{t-1} + g_{er,\hat{c}} \odot \hat{c} \\
    \end{aligned}
\end{equation}

Combining independent memories and shared memories, we can update cell state $c_t$ and hidden state $h_t$ for next time step as Equation \ref{c_t} and \ref{h_t}:

Meanwhile, we generate task specific independent representation $h_e, h_r$ and shared representation $h_{er}$ with corresponding memories:

\begin{equation}
    \left[\begin{array}{l}
    h_e \\
    h_s \\
    h_{er}
    \end{array}\right]=tanh(\left[\begin{array}{c}
    s_e \\
    s_s \\
    s_{er}
    \end{array}\right] )
\end{equation}

After we obtain specific representations for all tokens in the sentence, we leverage these representations to achieve entity extraction and relation extraction.

\paragraph{Difference with PFN~\cite{yan2021partition}}
Despite the similar objective to separate the task-specific and shared information in the token representations, SRN differs from PFN in the following aspects: 
(1) The \textit{selection} mechanism in SRN uses simple gates to select the relevant neurons to specific tasks. Whereas the \textit{partition} operation in PFN sorts neurons to aggregate neurons related to the same task so that PFN is only practicable for two-task learning. Instead, our \textit{selection} operation does not need to sort the neurons, allowing the scalability of our model to multiple tasks without impairing the effectiveness.
(2) PFN adds the independent memory and shared memory to obtain the final task-specific representation. In contrast, SRN directly utilizes the independent memory and leaves the work to the downstream task models on how to utilize task-specific and shared information from the representations. This allows more flexible use of the SRN outputs. 
(3) Inspired by the LSTM model, we incorporate the forget gate and output gate. Compared to PFN, forget gates allows our model to learn the forget mechanism. Besides, in PFN, the hidden state is just the result of cell state through a tanh function, while we add an output gate that enhances the expressing ability of our model.

\section{Analysis of Encoder}

\begin{table}[htbp]
\setlength{\tabcolsep}{4mm}
\caption{Comparison results of PFN and SRN under a two-task joint framework. }
  \centering
    \begin{tabular}{c|cc}
    \hline
    Model & Entity & Relation \\
    \hline
    PFN & 66.8 & \textbf{38.4} \\
    Table-SRN  & \textbf{67.8} & 38.3 \\
    \hline
    \end{tabular}%
  \label{tab:encoder}%
\end{table}%

SRN has a similar function as PFN \cite{yan2021partition} encoder, the main difference is that PFN uses a partition operation while SRN uses a selection operation 
The flexibility of the selection operation enables SRN to encode more than two sub-tasks. We perform experiments to check if replacing the PFN with SRN hurts the performance under the situation with two sub-tasks. We develop an SRN-Table model which only replaces PFN encoder of the joint table-filling framework designed by \citet{yan2021partition}. As shown in Table \ref{tab:encoder}, under the same entity-relation extraction framework, the model using SRN encoder increases the NER F1 score by 1\% and has similar RE performance to PFN. The results demonstrate SRN not only has better scalability but also improves the original encoding ability.